# QuotationFinder – Searching for Quotations and Allusions in Greek and Latin Texts and Establishing the Degree to Which a Quotation or Allusion Matches Its Source

Luc Herren

University of Münster, Germany – LHerren@uni-muenster.de

**Abstract**
The software programs generally used with the TLG (Thesaurus Linguae Graecae) [http1] and the CLCLT (CETEDOC Library of Christian Latin Texts) [http2] CD-ROMs are not well suited for finding quotations and allusions. QuotationFinder [http3] uses more sophisticated criteria as it ranks search results based on how closely they match the source text, listing search results with literal quotations first and loose verbal parallels last.

## INTRODUCTION
When looking for Greek and Latin quotations or allusions using standard search functions, a search with a Boolean "or" would yield too many matches. A search with a Boolean "and," however, would yield too few, as ancient authors often left out words when quoting without us having a chance to know in advance which ones they would keep. QuotationFinder produces better search results by using five criteria for determining if a given text is a quotation or an allusion: number and forms of words matched, their proximity to each other, how rare they are, and in what sequence they occur.

## I QUOTING IN ANTIQUITY

Designing software algorithms for finding quotations and allusions is not as straightforward as it may seem. In antiquity, authors did not use quotation marks or italics (though copyists or editors may have added these,) so that there is no simple obvious token in the text that search software could work with. In some contexts, one single word that the target and the source text have in common may suffice to constitute an allusion – for example if the context in the target names the source text, its author, the character originally uttering the word, or a phrase like "as it is written" (cf. [Schwerdtner, 2015] p. 26-27, [Plett, 1991] p. 9.) Authors may quote verbatim or they may assimilate the material to the target context (cf. [Plett 1988] p. 70-71.) Variations regarding syntax and semantics may occur – omissions, additions, changes in word forms, substitutions of words, etc. (cf. [Plett, 1991] p. 9.)
Formally, quotations vary greatly from each other (cf. [Schwerdtner, 2015] p. 25.39.) Quotation search engines should not be rigid in the way they look for formal features typical of quotations. The best approach seems to be to establish a score for every feature and rank potential quotations and allusions based on the sum of the scores they achieve regarding each of the features (cf. [Tischer, 2010] p. 103-106.)

# II RANKING PROSPECTIVE QUOTATIONS AND ALLUSIONS

## 2.1 Quantity
QuotationFinder establishes a score for every potential quotation or allusion as the text entered in the search form is matched against the collected ancient texts (the contents of the TLG or CLCLT.) The first parameter QuotationFinder uses is quantitative: Every word in the source (search text entered) gets points as it is matched in the target (set number of lines contained in the TLG/CLCLT.) If a word occurs n times in the source, it gets points n times as it is matched in the target. This means that if the word occurs only once in the source, but multiple times in the target, it still only gets points once (otherwise, special texts like entries in Suda, the 10$^{th}$ century encyclopedia, would get too many points.)

## 2.2 Quality
When ranking matches, QuotationFinder awards 3 points when the exact form of the word in the source text (entered in the search form) is found in the target (TLG/CLCLT). If there is a match for the same word, but in a different form (case, number, tense, etc.), 2 points are awarded. 1 point is given when the match is not the same word, but is derived from the same root or stem, or a synonym. (User enters roots, stems, synonyms in dedicated search fields.)

## 2.3 Rarity
If a rare word is matched in the target text, it is much more likely that we have found a quotation than if a ubiquitous word has been matched. Comparing the relative rarity of the word matched, QuotationFinder awards between 0 and 1 point per match. (User enters frequency information provided by TLG/CLCLT.)

## 2.4 Density
If a target paragraph contains more than one match, but the words matched are wide apart in our source text, we are far less likely to have found a quotation than if the matched words originally occur one after the other. This becomes more relevant the longer the search text is. The number of points a match is awarded for density is 1 divided by the number of words inserted between matched words plus 1.

## 2.5 Order
In the target text, the more the sequence of matched words resembles the sequence of these words in the source text, the higher the target text is to move up in the ranking. The number of points awarded per matched word is 1 divided by the number of words the match has "strayed" from its position in the source text plus 1.

## 2.6 Further Remarks
The density and order calculations are not trivial. As we can never know in advance which ones of our source text's words are going to be matched in the target text (as the author quoting may leave out any of them,) we cannot use any of their positions as reference points. So, QuotationFinder has to establish a "core" of the potential quotation first in order to gain a basis for these calculations. It does so by determining where in the target text the position of the matched words relative to each other most resembles the relative position of these words in the search text.
During development, by a process of trial and error, the relation of the points awarded for quantity, quality, rarity, density, and order was fine-tuned so as to establish a convincing rank order from exact quotation to loose verbal parallel. It proved useful to add up the points for quantity, quality and rarity, divide the sum by 3, then add the points for density multiplied times 2, and finally add the points for order.

QuotationFinder was no longer updated after 2008 as the TLG ceased to be published on CD-ROM, becoming an online only resource in a different data format ([http1] — the CLCLT is now available as part of the LLT DVD [http2].) Some of the ideas presented in this article may be of use as new software searching for quotations and allusions is developed. The QuotationFinder code (Perl CGI) is available on GitHub [http4].

**Conclusion**

In antiquity, people were not as strict as modern scholars in terms of the precision with which one was expected to follow the source when quoting (cf. [Schwerdtner, 2015], p. 34.) Classical authors may or may not provide explicit markers in their texts helping readers to note that they are quoting from or alluding to prior texts. When designing search software, it seems best to take into account a wide range of features indicative of quotations and allusions — like the number and forms of words matched, their proximity to each other, how rare they are, and in what sequence they occur, as was done with QuotationFinder.